\pdfoutput=1

\documentclass[11pt]{article}

\usepackage[]{emnlp2022}

\usepackage{times}
\usepackage{latexsym}

\usepackage[T1]{fontenc}

\usepackage[utf8]{inputenc}

\usepackage{microtype}

\usepackage{multirow}
\usepackage{booktabs}
\usepackage{makecell}

\usepackage{graphicx}

\usepackage{natbib}
\usepackage{amsfonts}
\usepackage{soul}
\usepackage{xspace}
\usepackage{amsmath}
\usepackage{longtable}
\newcommand{\ie}{\emph{i.e.,}\xspace}
\newcommand{\eg}{\emph{e.g.,}\xspace}

\usepackage{enumitem}
\setlist[itemize]{itemsep=-0.25ex,leftmargin=2.5ex}
\setlist[enumerate]{itemsep=-0.25ex,leftmargin=2.5ex}

\title{History-Aware Hierarchical Transformer for Multi-session Open-domain Dialogue System}
\author{Tong Zhang$^1$, Yong Liu$^{2,3}$, Boyang Li$^1$, Zhiwei Zeng$^3$, Pengwei Wang$^4$, Yuan You$^4$\\{\bf Chunyan Miao}$^{1,2,3}$, {\bf Lizhen Cui}$^5$\\
    $^1$School of Computer Science and Engineering, Nanyang Technological University, Singapore\\
    $^2$Alibaba-NTU Singapore Joint Research Institute, Nanyang Technological University, Singapore\\
    $^3$Joint NTU-UBC LILY Research Centre, Nanyang Technological University, Singapore\\
    $^4$Alibaba Group, China\\
    $^5$School of Software, Shandong University, China}

\begin{document}
\maketitle

\begin{abstract}
With the evolution of pre-trained language models, current open-domain dialogue systems have achieved great progress in conducting one-session conversations. In contrast, Multi-Session Conversation (MSC), which consists of multiple sessions over a long term with the same user, is under-investigated. In this paper, we propose History-Aware Hierarchical Transformer (HAHT) for multi-session open-domain dialogue. HAHT maintains a long-term memory of history conversations and utilizes history information to understand current conversation context and generate well-informed and context-relevant responses. Specifically, HAHT first encodes history conversation sessions hierarchically into a history memory. Then, HAHT leverages historical information to facilitate the understanding of the current conversation context by encoding the history memory together with the current context with attention-based mechanisms. Finally, to explicitly utilize historical information, HAHT uses a history-aware response generator that switches between a generic vocabulary and a history-aware vocabulary. 
Experimental results on a large-scale MSC dataset suggest that the proposed HAHT model consistently outperforms baseline models. Human evaluation results support that HAHT generates more human-like, context-relevant and history-relevant responses than baseline models.  
\end{abstract}

\section{Introduction}
Open-domain dialogue systems, also known as chatbots, are designed to converse with and engage users on any topic with the aim of establishing, maintaining, and strengthening long-term relationships \cite{clark2019makes, roller2020open}. 
Recently, open-domain dialogue systems built based on large-scale generative pre-trained models \cite{adiwardana2020towards,roller-etal-2021-recipes,zhang-etal-2020-dialogpt} have substantially improved the performance of chatbots.

However, most existing chatbots are designed to interact with users in a single conversation session. When the current session ends, the chatbot forgets its contents and will commence a new independent session with the same user next time. 
When previously discussed topics reemerge, such chatbots often appear ignorant and fail to reengage users appropriately. The apparent forgetfulness limits the chatbots’ ability to establish and maintain long-term relationships with users.

\begin{figure}[t]
    \centering
    \includegraphics[width=\columnwidth]{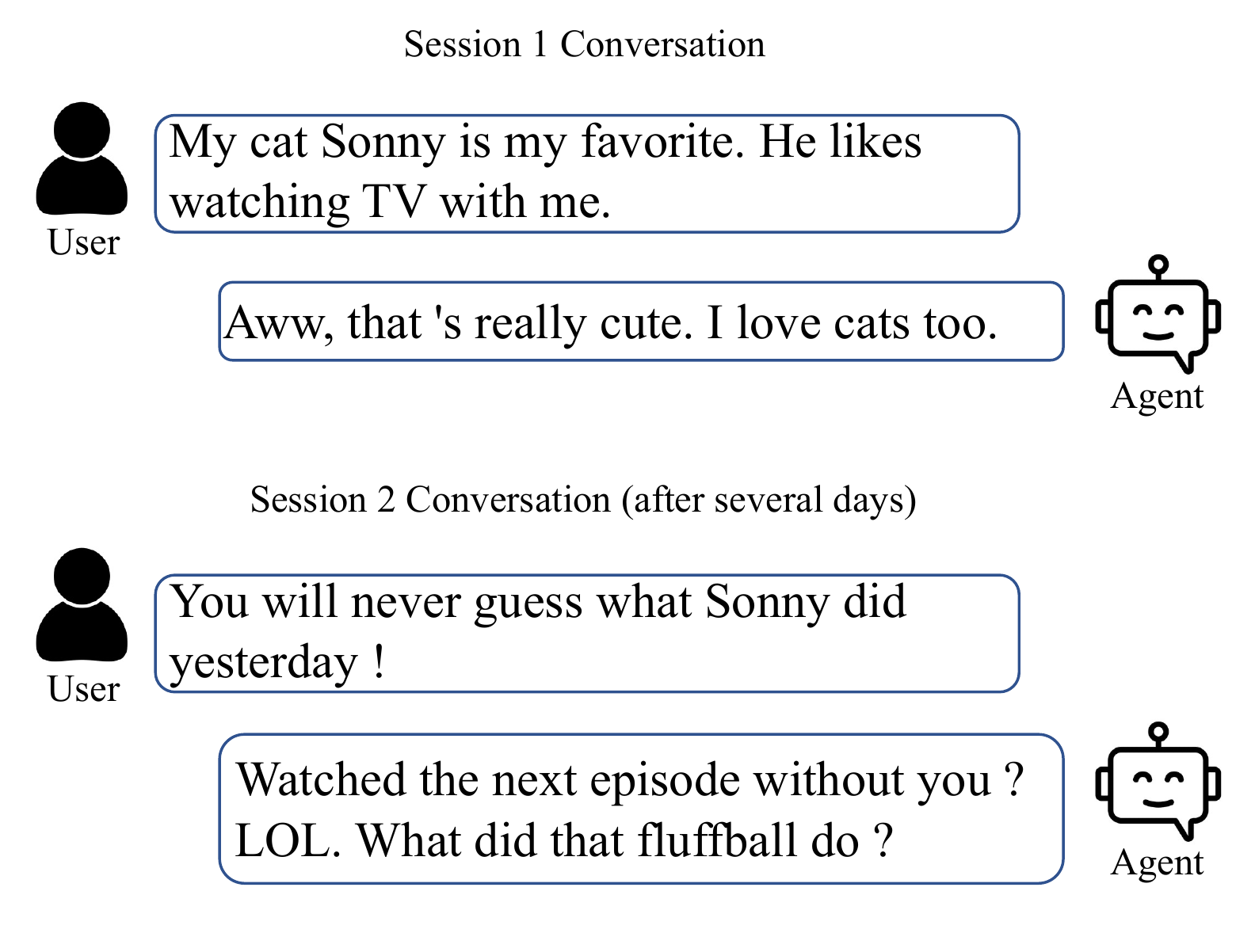}
    \caption{An illustrated example of a two-session conversation between a user and an agent.}
    \label{fig:conv_example}
\end{figure}


We argue that, to better engage users in multi-session conversations (MSCs), a chatbot should maintain a long-term memory of historical contexts, which allows the chatbot to reengage the user appropriately when similar contexts reemerge. 
By learning from historical conversations, the chatbot should gradually refine its understanding of and deepen its relationship with the user. Figure~\ref{fig:conv_example} shows an example of a two-session conversation between a user and a chatbot. In the second session, the chatbot infers  that Sonny is a cat and generates the response based on the history information that Sonny likes watching TV with the user.

History-aware chatbots will be able to generate more well-informed and context-relevant responses, which can help to elicit long-term commitments and develop emotional attachments from users to sustain close relationships over time. To this end, we propose the History-Aware Hierarchical Transformer (HAHT) for multi-session open-domain dialogue systems, which can effectively leverage history conversations to conduct more engaging MSCs. 
HAHT maintains a long-term memory to store historical conversational contexts, which is updated when a new session is conducted. Based on the long-term memory and the context in the current session, relevant tokens in historical contexts are selected to adapt the current response.

Specifically, as the number of tokens in a conversation utterance and the number of turns in a conversation are usually not very long\footnote{On average, conversations have 13 turns and conversation utterances have 16 tokens in Facebook MSC dataset.}, we first encode the history conversation hierarchically into the history memory using Transformer \cite{vaswani2017attention}. The history memory serves as a high-level representation of history conversations. Secondly, as history conversations usually can facilitate the understanding of the current conversation context, we design a history-aware context encoder. The context encoder encodes conversation context, considering both history conversations and the current conversation, by adopting the transformer attention over the history memory and current conversation context. Then, the context encoder also updates the history memory based on the current conversation context. Finally, we design a history-aware decoder to fuse learned history information into the response generation process. The history-aware decoder can switch between two strategies, \ie generating a word from the generic vocabulary or directly copying a word from history conversations. 

Experimental results on the large-scale Facebook MSC dataset show that the proposed HAHT model outperforms previous multi-session open-domain dialogue systems in various evaluation metrics. Human evaluation results support that HAHT generates more readable, context-relevant, and history-relevant responses than baseline models. In addition, the ablation study confirms that both the hierarchical encoding of history conversations and the history-aware decoder contribute greatly to HAHT's performance on MSCs and help it leverage historical information more effectively.

\section{Related Work}
Open-domain dialogue systems aim to perform chit-chat without task and domain restrictions \cite{ritter-etal-2011-data} and establish long-term relationships with users \cite{clark2019makes, roller2020open}. They are generally divided into two groups: generation-based systems and retrieval-based systems. Retrieval-based systems seek to find a suitable response from a large response candidate set \cite{zhou-etal-2016-multi,yuan-etal-2019-multi, zhong-etal-2020-towards, retrieval-zhou-2021, qian-cikm-2021}, whereas, generation-based systems focus on generating responses from scratch based on the dialogue history \cite{serban2016building, xiaoice-2018, adiwardana2020towards, roller2020open, xu2021beyond}. In this paper, we focus on generation-based systems. 

Early approaches to response generation include template-based generation methods \cite{higashinaka-etal-2014-towards} and statistical machine translation (SMT) methods \cite{ritter-etal-2011-data}. With the development of deep learning, sequence-to-sequence (Seq2seq) models have been applied to generation-based dialogue systems and achieved great performance \cite{li-etal-2016-diversity, vinyals2015neural, VHRED2016}. Recently, with the increasing availability of large-scale dialogue datasets \cite{li-etal-2017-dailydialog, zhang-etal-2018-personachat, dinan2018wizard, huang2020challenges}, Transformer-based language models pretrained with large-scale corpora, such as Meena \cite{adiwardana2020towards}, BlenderBot \cite{roller-etal-2021-recipes}, DialogueGPT \cite{zhang-etal-2020-dialogpt}, and PLATO \cite{platonov-etal-2020-spoken}, have made significant progress in the area of open-domain dialogues. 

Despite the advancements in the field, current state-of-the-art generative pre-trained models are designed for and trained on large datasets of single-session conversations with a small number of turns. As a result, most existing models employ short token truncation lengths, such as 128 tokens for Meena \cite{adiwardana2020towards}, and are unable to encode and utilize historical contexts in MSCs effectively. In addition, there is also a lack of public MSC datasets.
\citeauthor{xu2021beyond} released the first multi-session conversation dataset, \ie \textit{Facebook MULTI-SESSION CHAT (Facebook MSC)}, and explored different retrieval-augmented generative models on the dataset \cite{lewis2020retrieval, shuster-etal-2021-retrieval-augmentation}, which achieved better results than the standard Transformer \cite{vaswani2017attention}. 
However, the experimental results demonstrate that their methods need to retrieve a very large portion of history conversations to achieve better results than the standard Transformer. In addition, these models still need to concatenate the retrieved raw history conversation text with the current conversation context, yielding concatenations that are still much longer than the 128 token truncation lengths. Therefore, the incorporation of historical contexts in these methods is still limited by the short token truncation lengths of pre-trained models.


\begin{figure}
    \centering
    \includegraphics[width=\columnwidth]{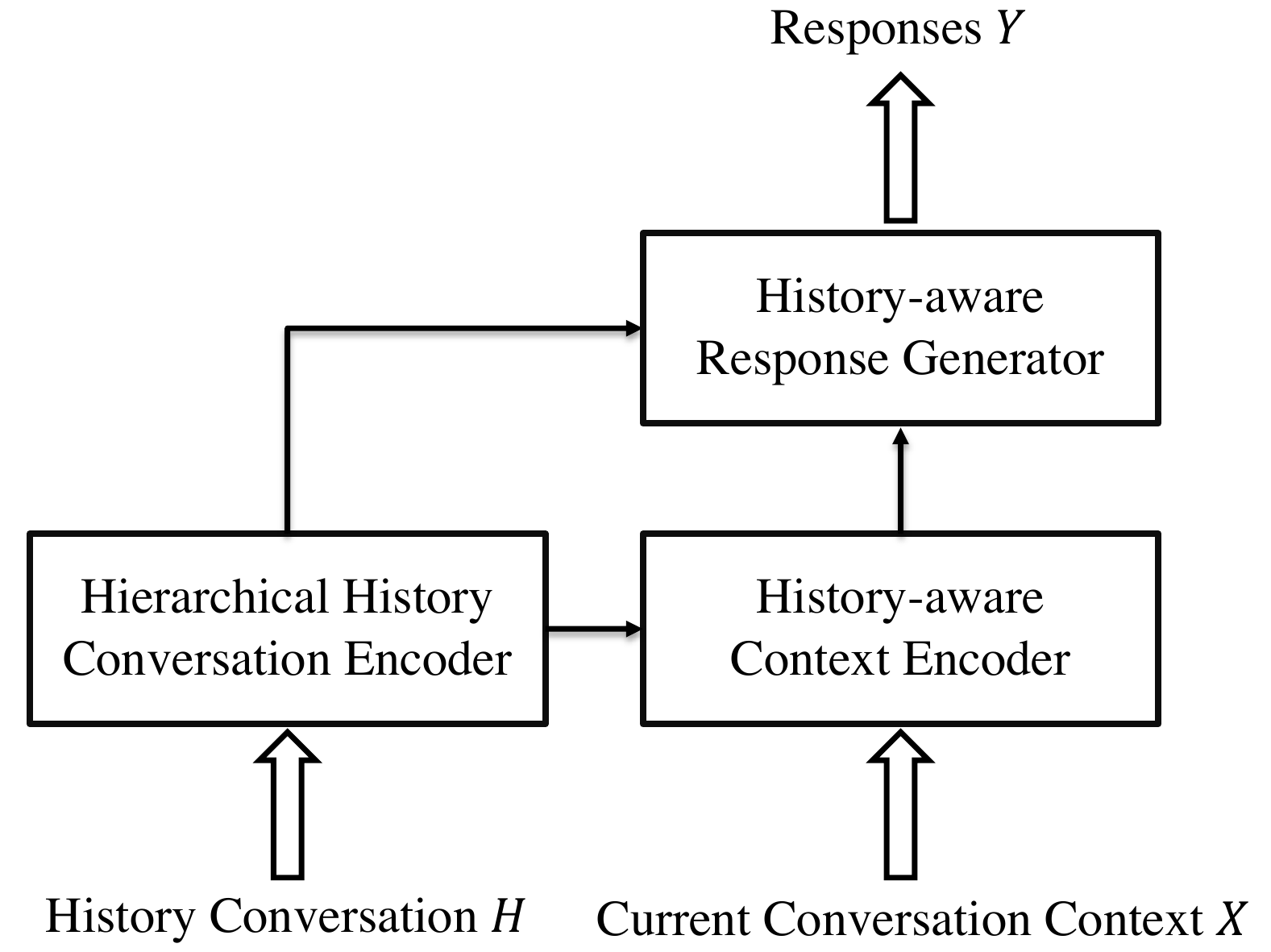}
    \caption{The overall structure of the proposed HAHT model, which contains 1) hierarchical history conversation encoder, 2) history-aware context encoder, and 3) history-aware response generator. The details of each component are shown in Figure \ref{fig:hist_encoder}, \ref{fig:ctx_encoder}, \ref{fig:decoder}, respectively.}
    \label{fig:overall_structure}
\end{figure}

\section{The Proposed Method}

In general, a \textit{Multi-Session Conversation (MSC)} consists of a current conversation session and several history conversation sessions that happen before the current one, all between the same two interlocutors. A multi-session open-domain dialogue system aims to generate natural, well-informed, and context-relevant responses to the user's utterances based on all history conversation sessions and the current conversation context.

Formally, we denote the MSC dataset $D$ by a list of $N$ conversations in the format of ($H, X, y$). Here, $X = \{x_1, x_2, \cdots, x_{n_x}\}$ denotes $n_x$ context utterances of the current conversation session. $H=\{H^1, H^2, \cdots, H^{M}\}$ denotes $M$ history conversation sessions, where $H^i = \{h^i_1, h^i_2, \cdots, h^i_{n_i}\}$ denotes $n_i$ chronologically ordered utterances of the $i$-th history conversation session.  
$y$ is the ground truth response to $X$ under the background of $H$. The MSC task can be formulated as learning a function $f(H, X)$ to predict the next utterance $x_{n_x+1}$ based on $H$ and $X$.

In this work, we propose a novel model, namely HAHT, for the MSC task. Figure~\ref{fig:overall_structure} shows the overall structure of HAHT, which consists of three main components: 1) hierarchical history conversation encoder, 2) history-aware context encoder, and 3) history-aware response generator. We present the details of each component of HAHT as follows.

\begin{figure}
    \centering
    \includegraphics[width=\columnwidth]{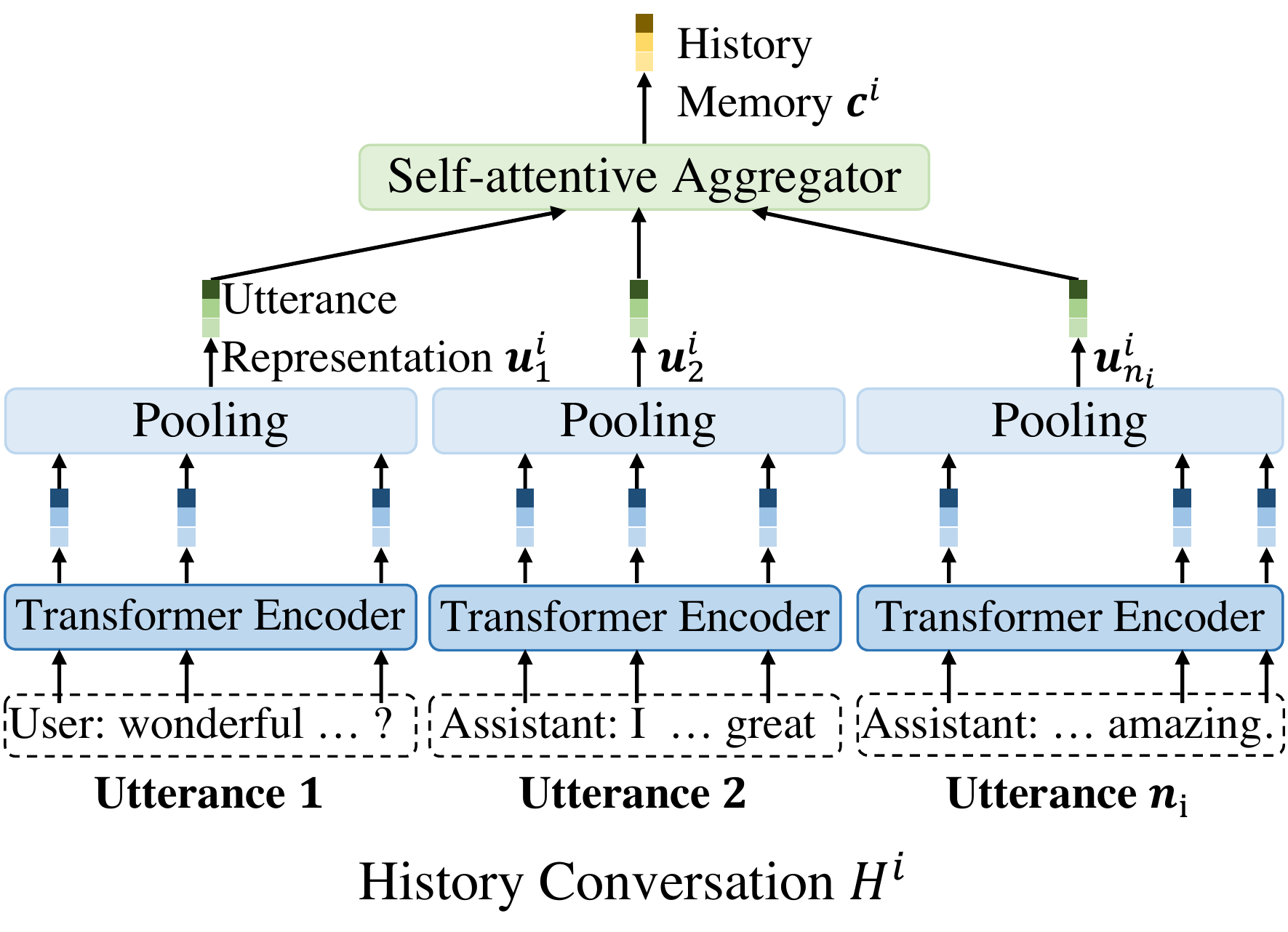}
    \caption{The structure of the hierarchical history conversation encoder in HAHT.}
    \label{fig:hist_encoder}
\end{figure}

\subsection{Hierarchical History Conversation Encoder}
The main challenge in encoding history conversation sessions is the limited maximum input length imposed by pre-trained dialogue systems. 
If all history conversations are simply concatenated and fed into the pre-trained dialogue system, the length of the concatenation will exceed the maximum input length. Thus, most parts of the input will be truncated.
To preserve more information in the history conversation, we encode each history conversation session separately in a hierarchical fashion.

Specifically, for a history conversation session $H^i = \{h^i_1, h^i_2, \cdots, h^i_{n_i}\}$, we first prepend a special token ``User:'' or ``Assistant:'' to each utterance $h^i_j$ in $H^i$ depending on the role of the utterance speaker, and then pad all utterances to the same length $l_{utter}$. For each utterance $h^i_j$, we apply an embedding layer $E_m$, $n_{enc} $ Transformer encoder layers, and a Max-pooling layer to obtain its dense representation as follows,
\begin{equation}
    \mathbf{u}_j^i = \text{Max-pooling}\big(\text{Transformer}_{n_{enc}}(E_m(h_j^i))\big),
\end{equation}
where $\mathbf{u}_j^i \in \mathbb{R}^d$. Moreover, we denote all the utterance representations in the history conversation $H^i$ by $\mathbf{U}^i = \{\mathbf{u}_1^i, \mathbf{u}_2^i, \cdots, \mathbf{u}_{n_i}^i\}\in \mathbb{R}^{n_i\times d}$ , where $n_i$ is the turn number of $H^i$.
Next, we apply a conversation aggregator $F_c$ to aggregate all utterance representations $\mathbf{U}^i$ into the condensed history memory $\mathbf{c}^i$,
\begin{equation}
    \mathbf{c}^i = F_c(\mathbf{U}^i).
\end{equation}
The conversation aggregator is developed based on the following self-attentive mechanism~\cite{lin2017iclr}, 
\begin{align}
        &F_c(\mathbf{U}^i) = \mathbf{\alpha} \mathbf{U}^i, \nonumber\\
        & \mathbf{\alpha} = \mbox{softmax}\big(\mathbf{W}_k\mbox{tanh}(\mathbf{W}_q\mathbf{U}^{i\top})\big),
\end{align}
where $\mathbf{W}_q$ and $\mathbf{W}_k$ are learnable parameters. $\mathbf{\alpha} \in \mathbb{R}^{n_i}$ is the importance vector of the history conversation utterances in $H^i$.

After applying previous steps to all history conversations $H$, we will finally obtain a history memory matrix $\mathbf{C} \in \mathbb{R}^{M\times d}$ containing a history memory for each history conversation, where $M$ is the number of history conversation sessions.

\begin{figure}
    \centering
    \includegraphics[width=\columnwidth]{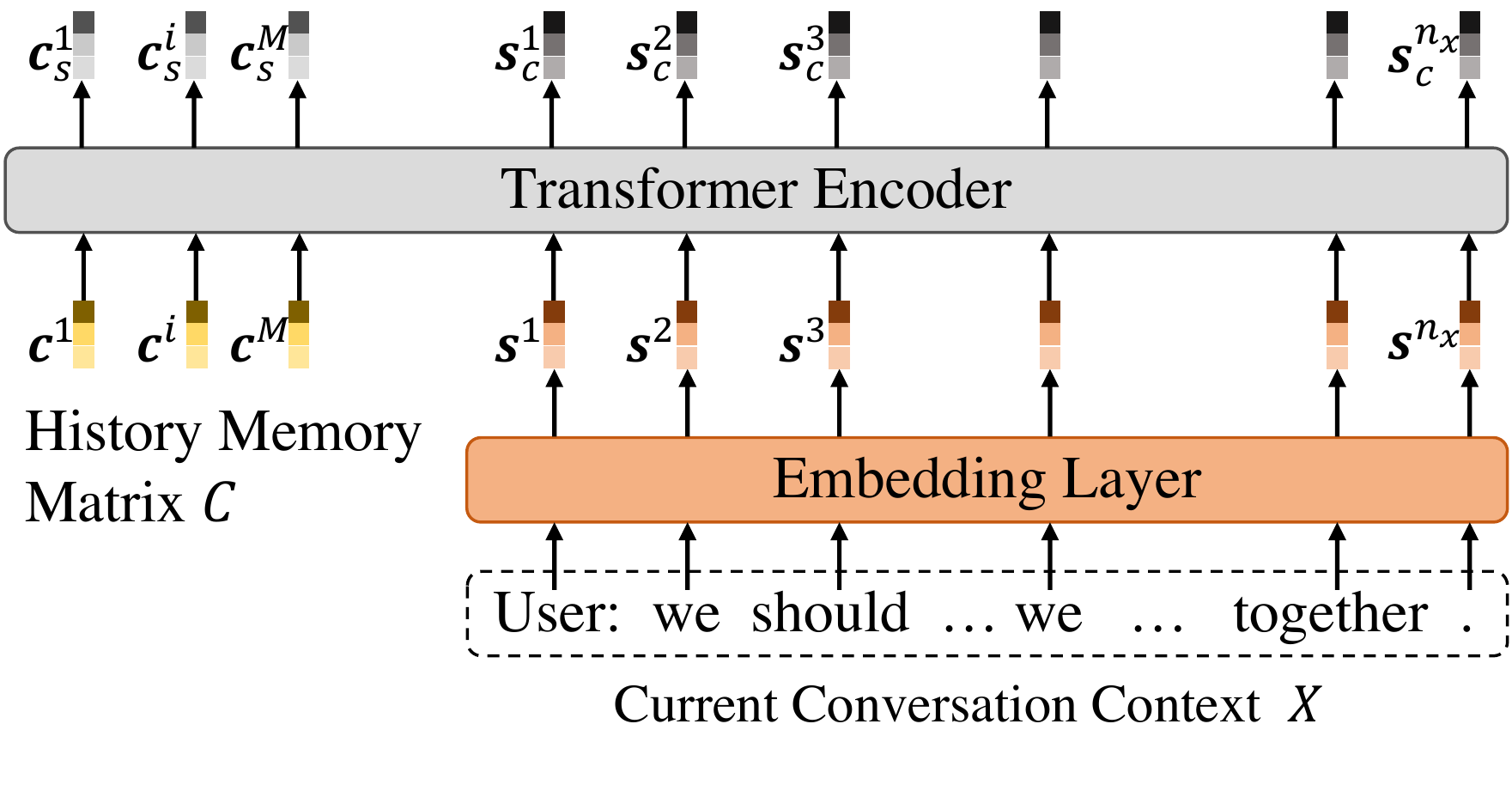}
    \caption{The structure of the history-aware context encoder in HAHT.}
    \label{fig:ctx_encoder}
\end{figure}

\subsection{History-aware Context Encoder}
History conversation sessions usually contain the background stories (\eg interlocutors' profiles or previous discussions between them) that bring out the current conversation session. Leveraging the history conversations will help the model to better understand the current conversation context and respond properly. On the other hand, the current conversation context can help the model update the history memories.
Thus, we encode the history memory $\mathbf{C}$ together with the current conversation context by adopting the transformer attention between them.

For the current conversation context $X$, we also prepend a special token ``User:'' or ``Assistant:'' to each utterance depending on the role of the utterance speaker and concatenate all utterances into a single sentence. Then, we adopt the embedding layer $E_m$ to obtain a sequence of context token embeddings $\mathbf{S} = \{\mathbf{s}^1, \mathbf{s}^2, \cdots,\mathbf{s}^{n_x}\}$, where $n_x$ is the length of the context sequence. Next, we concatenate the history memory matrix $\mathbf{C}\in\mathbb{R}^{M\times d}$ with $\mathbf{S}\in\mathbb{R}^{n_x\times d}$ over the first dimension and apply $n_{enc}$ Transformer encoder layers. 

By employing attention in the transformer encoder layers, our model can understand the conversation context by attending to all context token embeddings and history conversation memories. We denote this history-aware context encoding by $\mathbf{S}_c \in \mathbb{R}^{n_x\times d}$.  After context encoding, history conversation memories are updated based on the latest information from the current conversation context. We denote this context-updated history memory as $\mathbf{C}_s \in \mathbb{R}^{M\times d}$. The concatenation of $\mathbf{C}_s$ and $\mathbf{S}_c$ over the first dimension will become the input of the response generator.

\begin{figure}
    \centering
    \includegraphics[width=\columnwidth]{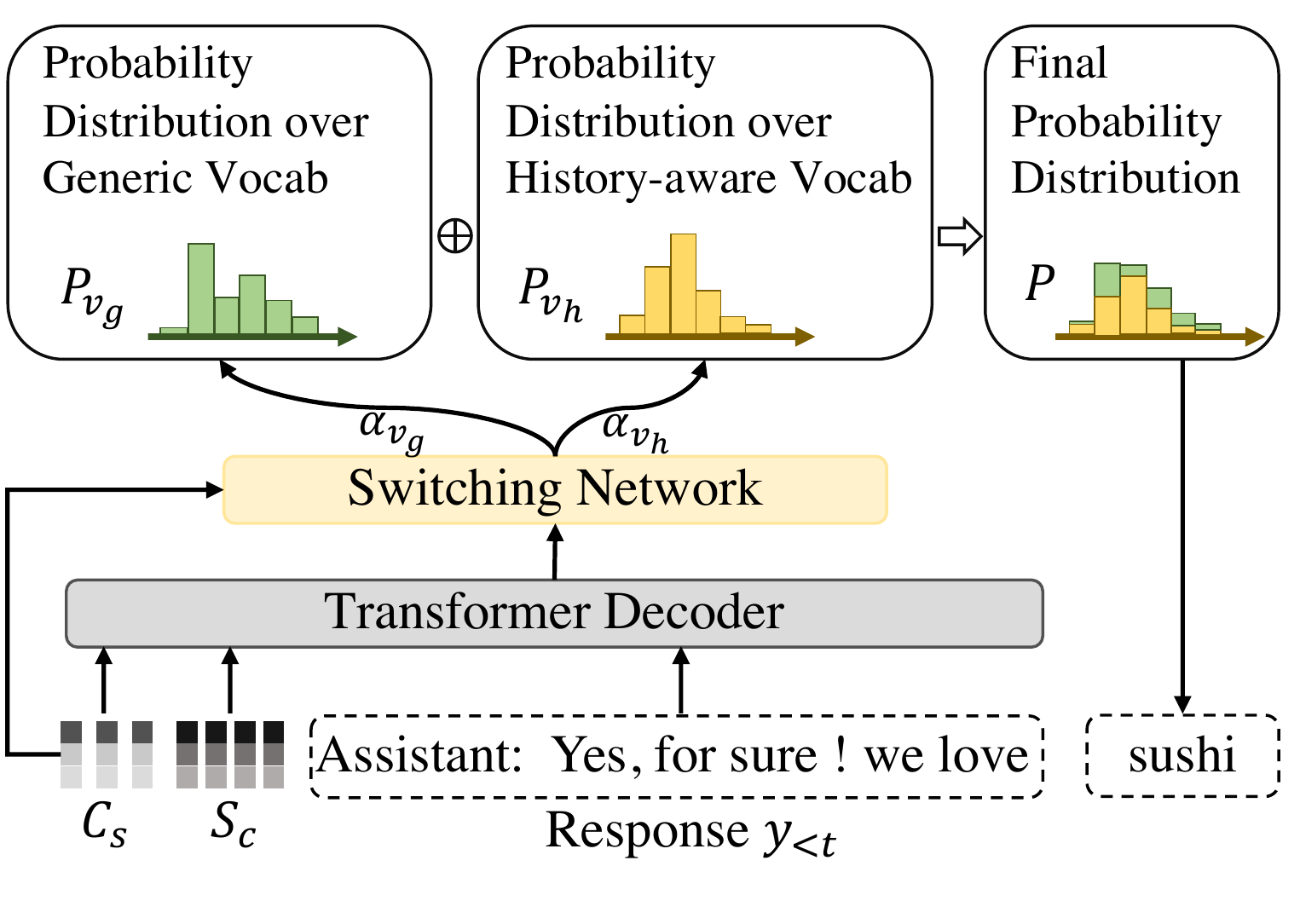}
    \caption{The structure of the history-aware response generator in HAHT.}
    \label{fig:decoder}
\end{figure}

\subsection{History-aware Response Generator}
Inspired by CopyNet~\cite{gu-etal-2016-incorporating}, we construct two vocabularies, \ie generic vocabulary $V_g$ and history-aware vocabulary $V_h$, to better generate history-aware responses. The generic vocabulary $V_g$  contains the words that appear in all the training dataset, and the history-aware vocabulary $V_h$ only contain the words that appear in the history conversations $H$. To generate a word of the response, the response generator will choose to generate a generic word from $V_g$ or directly copy a word from $V_h$ based on the switching mechanism~\cite{gulcehre-etal-2016-pointing}.

Specifically, at each decoding time step $t$, we feed $\mathbf{C}_s$ , $\mathbf{S}_c$ and the ground truth word sequence before $t$ into $n_{dec}$ Transformer decoder layers and obtain a hidden representation vector $\mathbf{o}_t \in \mathbb{R}^d$. The probability distribution over the generic vocabulary $V_{g}$ at the decoding time step $t$ is computed as,
\begin{equation}
    P_{v_g} = \mbox{softmax}\big(\mbox{FC}_1(\mathbf{o}_t)\big),
\end{equation}
where $\mbox{FC}_1$ is a fully connected layer.

To calculate the probability distribution over the history-aware vocabulary $V_{h}$, we adopt a max-pooling layer over the context-updated history memory $\mathbf{C}_s$, a fully connected layer, and a softmax function as follows,
\begin{equation}
    P_{v_h} = \mbox{softmax}\big(\mbox{FC}_2(\mbox{max-pooling}(\mathbf{C}_s))\big),
\end{equation}
where $\mbox{FC}_2$ is a fully connected layer.

The final word probability distribution at time step $t$ is computed by using a switching mechanism between $P_{v_g}$ and $P_{v_h}$ as follows,
\begin{equation}
    P = \alpha_{v_g} * P_{v_g} + \alpha_{v_h} * P_{v_h},
\end{equation}
where $ \alpha_{v_g}$ and $\alpha_{v_h}$ is the switching probability of generating from generic vocabulary or copying from history conversations.
$ \alpha_{v_g}$ and $ \alpha_{v_h}$ is calculated as follows,
\begin{align}
    [\alpha_{v_g}, & ~\alpha_{v_h}]  = \nonumber\\
    &\mbox{softmax}\big(\mbox{FC}_3([o_j;\mbox{max-pooling}(\mathbf{C}_s)])\big),
\end{align}
where $\mbox{FC}_3$ is a fully connected layer, and [;] is a concatenation operation over the last dimension.

\subsection{Model Training}
We train the model to maximize the generation probability of the target response, given the current conversation context and history conversations in an end-to-end manner. The loss function of HAHT is defined as,
\begin{equation}
    \mathcal{L} = -\sum_{t=1}^{n_y}\log\big(P(y_j|X, H, y_{<t})\big),
\end{equation}
where $X$ denotes the current conversation context, $H$ denotes all history conversations, $y_{<t}$ denotes tokens before time step $t$, and $n_y$ denotes the length of the ground truth response.

\begin{table}[t]
    \centering
    \resizebox{\linewidth}{!}{
    \begin{tabular}{c|cc|cc|cc}
        \toprule[1pt]
         \multirow{2}*{\makecell{Session\\number}} & \multicolumn{2}{c|}{Train} & \multicolumn{2}{c|}{Valid} & \multicolumn{2}{c}{Test} \\
          & Conv. & Utter. & Conv. & Utter. & Conv. & Utter. \\
         \hline
         1$^{\star}$ & 8939 & 131,438 & 1000 & 7,801 & 1015 & 6,634\\
         2 & 4000 & 46,420 & 500 & 5,897 & 501 & 5,939\\
         3 & 4000 & 47,259 & 500 & 5,890 & 501 & 5,924\\
         4 & 1001 & 11,870 & 500 & 5,904 & 501 & 5,940\\
         5 & - & - & 500 & 5,964 & 501 & 5,945\\
         \hline
         Total & - & 236,987 & - & 31,456 & - & 30,382\\
        \toprule[1pt]
    \end{tabular}
}
    \caption{The statistics of Facebook Multi-Session Chat (Facebook MSC) Dataset. Session number $i$ indicates there are $i\textnormal{-}1$ history conversation sessions that happen before the last conversation session. $^{\star}$: Session 1 does not contain history conversation sessions.}
    \label{tab:facebook_msc}
\end{table}

\begin{table*}[t]
    \centering
    \resizebox{\textwidth}{!}{
    \begin{tabular}{c|c|c|c|c|c|c|c|c|c|c|c|c}
    \hline
        \multirow{2}*{Model} & \multicolumn{3}{c|}{Session 2} & \multicolumn{3}{c|}{Session 3} & \multicolumn{3}{c|}{Session 4} & \multicolumn{3}{c}{Session 5} \\
    \cline{2-13}
        & B-2 & B-3& R-L& B-2 & B-3& R-L & B-2 & B-3& R-L & B-2 & B-3& R-L \\
    \hline
        BlenderBot & 2.79 & 0.65 & 13.73 & 2.41 & 0.45 & 13.06 & 2.14 & 0.39 & 12.76 & 2.26 & 0.45 & 12.75\\
        BlenderBot$_{\text{msc}}$ & 4.76 & 1.51 & 16.18 & 5.03 & 1.61 & 16.39 & 4.78 & 1.49 & 15.56 & 4.98 & 1.48 & 16.10 \\
        FID-RAG & 4.82 & 1.54 & 16.53 & 5.04 & 1.61 & 16.42 & 4.84 & 1.48 & 15.89 & 5.06 & 1.57 & 16.01 \\
        HAHT (ours) & \textbf{5.07} & \textbf{1.57} & \textbf{16.90} & \textbf{5.27} & \textbf{1.67} & \textbf{16.72} & \textbf{5.00} & \textbf{1.55} & \textbf{15.97} & \textbf{5.16} & \textbf{1.60} & \textbf{16.42}\\

    \hline
    \end{tabular}
    }
    \caption{Automatic evaluation results of different models on all session data. Session $i$ indicates there are $i\textnormal{-}1$ history conversation sessions. B-2, B-3, and R-L denote BLEU-2, BLEU-3, and Rouge-L respectively. The best results are in \textbf{boldface}.
    }
    \label{all_session_results_table}
\end{table*}

\section{Experimental Settings}
In this section, we introduce the experimental dataset, evaluation metrics, baseline methods, and model settings.

\subsection{Experimental Dataset}

The experiments are performed on a large dataset, \ie Facebook MULTI-SESSION CHAT (Facebook MSC)~\cite{xu2021beyond}. It is a crowdsourced dataset consisting of multi-session conversations, where the interlocutors learn about each other's interests and discuss the things they have understood from past sessions. The number of history conversations in Facebook MSC varies from 1 to 4. Session number $i$ indicates there are $i\textnormal{-}1$ history conversations happening before the last conversation session.
The statistics of the Facebook MSC dataset are summarized in Table \ref{tab:facebook_msc}. As session 1 does not have history conversations, we evaluate our model on session 2-5.

\begin{table}
    \centering
    \resizebox{\linewidth}{!}{
	\begin{tabular}{c|c|c|c}
    \hline
     Model & Readability & \makecell{Context \\ Relevancy} & \makecell{History \\ Relevancy} \\
    \hline
        BlenderBot & 1.78 & 1.13 & 0.09\\
    \hline
        BlenderBot$_\text{msc}$ & 1.82 & 1.56  & 0.13\\
    \hline
        RAG-FID & 1.89 & 1.84 & 0.21\\
    \hline
        HAHT (ours) & \textbf{2.05} & \textbf{2.03} & \textbf{0.33}\\
    \hline
    \end{tabular}}
    \caption{Human evaluation of the response generation by different methods. 
    All scores are rated in four levels $0$/$1$/$2$/$3$. The best results are in \textbf{boldface}. We measure the inter-rater reliability with Fleiss’ Kappa \cite{kappa}. Our annotations obtain “good agreement” for Readability (0.614) and “moderate agreement” for Context Relevancy (0.526) and History Relevancy (0.573).}
    \label{response_huamn_table}
\end{table}

\subsection{Evaluation Metrics}
We conduct both automatic and human evaluations to demonstrate the effectiveness of the proposed model. For automatic evaluations, we leverage BLEU-2, BLEU-3 \cite{papineni-etal-2002-bleu}, and ROUGE-L \cite{lin-och-2004-automatic} to measure word overlaps between the generated response text and ground truth text. 

Moreover, we also randomly sample 50 MSCs from the test set to conduct human evaluations. We present all the history conversation sessions, current conversation context, and the generated responses to three well-educated annotators. The annotators will evaluate the quality of the generated responses from the following three aspects: 
\begin{itemize}
    \item \textbf{Readability}: measures whether the generated responses are natural and fluent.
    \item \textbf{Context Relevancy}: measures whether the generated responses are correlated with the current conversation context.
    \item \textbf{History Relevancy}: measures whether the generated responses are correlated with history conversations. Only responses that are consistent with history conversations are considered relevant to history.
\end{itemize}

Each aspect is rated in four different levels 0/1/2/3, and the final score of each aspect is the average of the scores given by all annotators. We measure the inter-annotator reliability with Fleiss’ Kappa \cite{kappa}. For all evaluation metrics, the higher value indicates better performance.

\subsection{Baseline Methods}
We compare the proposed HAHT model with the following baseline methods.
\begin{itemize}
    \item \textbf{BlenderBot} \cite{roller-etal-2021-recipes}: This is a large-scale open-domain dialogue model pre-trained on the dialogue data scraped from social discussions on the web.
    \item \textbf{BlenderBot$_{\text{msc}}$}: This is the BlenderBot model finetuned on the MSC dataset.
    \item \textbf{FID-RAG} \cite{shuster-etal-2021-retrieval-augmentation}: In this method, RAG-trained retriever~\cite{lewis2020retrieval} is used to retrieve top-$N$ history conversations, and Fusion-Decoder (FiD) \cite{izacard-grave-2021-leveraging} is adopted to generate a final response considering the retrieved history conversations and current conversations. Following \cite{xu2021beyond}, $N$ is empirically set to 5.
\end{itemize}

\subsection{Model Settings}
In this work, all the evaluated methods are trained following the same settings.
Due to the limitation of computation resources, we use the BlenderBot model with 90M parameters as the initial pre-trained model and finetune it on the Facebook MSC dataset. The input length truncation is set to 256. The number of Transformer encoder layers $n_{enc}$ and decoder layers $n_{dec}$ are both set to 12. For model training, we use the Adamax optimizer~\cite{kingma2014adam} with a learning rate of $1\times10^{-6}$, batch size of 16, dropout ratio of 0.1, and early stopping patience of 10. All the fine-tuned models are trained with a maximum of two 32GB GPUs (NVIDIA V100).

\begin{table*}[t]
    \centering
    \resizebox{\textwidth}{!}{
    \begin{tabular}{c|c|c|c|c|c|c|c|c|c|c|c|c}
    \hline
         \multirow{2}*{Model} & \multicolumn{3}{c|}{Session 2} & \multicolumn{3}{c|}{Session 3} & \multicolumn{3}{c|}{Session 4} & \multicolumn{3}{c}{Session 5} \\
    \cline{2-13}
        & B-2 & B-3& R-L& B-2 & B-3& R-L& B-2 & B-3& R-L & B-2 & B-3& R-L \\
    \hline
        BlenderBot &4.71 & 1.47 & 18.20 & 3.85 & 0.93 & 17.10 & 3.69 & 0.83 & 16.78 & 4.00 & 1.19 & 17.19 \\
        BlenderBot$_{\text{msc}}$ & 6.39 & 2.56 & 19.30 & 5.82 & 1.93 & 18.67 & 5.30 & 1.76 & 17.9 & 6.10 & 2.30 & 18.65\\
        FID-RAG & 6.41 & 2.51 & 19.82 & 5.83 & 1.95 & 18.38 & \textbf{5.81} & 1.85 & \textbf{18.44} & 6.02 & 2.27 & 18.52 \\
        HAHT (ours) & \textbf{6.69} & \textbf{2.73} & \textbf{20.02} & \textbf{6.03} & \textbf{2.20} & \textbf{18.70} & 5.48 & \textbf{1.95} & 18.00 & \textbf{6.38} & \textbf{2.51} & \textbf{19.18}\\
    \hline
    \end{tabular}
    }
    \caption{Automatic evaluation results of different models on session-opening data. Session $i$ indicates there are $i\textnormal{-}1$ history conversation sessions. B-2, B-3, and R-L denote BLEU-2, BLEU-3, and Rouge-L respectively. The best results are in \textbf{boldface}.}
    \label{session_opening_results_table}
\end{table*}

\begin{table*}[t]
    \centering
    \resizebox{\textwidth}{!}{
    \begin{tabular}{c|c|c|c|c|c|c|c|c|c|c|c|c}
    \hline
        \multirow{2}*{Model} & \multicolumn{3}{c|}{Session 2} & \multicolumn{3}{c|}{Session 3} & \multicolumn{3}{c|}{Session 4} & \multicolumn{3}{c}{Session 5} \\
    \cline{2-13}
        & B-2 & B-3& R-L& B-2 & B-3& R-L & B-2 & B-3& R-L & B-2 & B-3& R-L \\
    \hline
        HAHT & \textbf{5.07} & \textbf{1.57} & \textbf{16.90} & \textbf{5.27} & \textbf{1.67} & \textbf{16.72} & \textbf{5.00} & \textbf{1.55} & \textbf{15.97} & \textbf{5.16} & \textbf{1.60} & \textbf{16.42}\\
        HAHT\textsubscript{w/o HIER} & 5.00 & 1.57 & 16.72 & 5.19 & 1.63 & 16.61 & 4.86 & 1.49 & 15.90 & 5.10 & 1.57 & 16.21\\
        HAHT\textsubscript{w/o HIST} & 4.98 & 1.50 & 16.81 & 5.09 & 1.58 & 16.51 & 4.75 & 1.45 & 15.51 & 5.10 & 1.49 & 16.24\\
        HAHT\textsubscript{w/o SW} & 5.01 & 1.56 & 16.86 & 5.19 & 1.61 & 16.46 & 4.87 & 1.55 & 15.88 & 5.07 & 1.55 & 16.17\\
    \hline
    \end{tabular}
    }
    \caption{The performance achieved by HAHT and different HAHT variants. Session $i$ indicates there are $i\textnormal{-}1$ history conversation sessions. B-2, B-3, and R-L denote BLEU-2, BLEU-3, and Rouge-L respectively. The best results are in \textbf{boldface}.}
    \label{ablation_study}
\end{table*}

\section{Experimental Results}
This section presents the experimental results of the automatic evaluation, human evaluation, evaluation on session openings, ablation study, and case study.

\subsection{Automatic Evaluation}

The automatic evaluation results of different models are shown in Table~\ref{all_session_results_table}. It can be observed that BlenderBot$_\text{msc}$ performs much better when finetuned on the MSC dataset. FID-RAG performs better than BlenderBot$_\text{msc}$. The potential reason is that RAG can retrieve important history conversations, and FID can combine the retrieved conversations with current conversations to generate better responses. Moreover, the proposed HAHT model consistently outperforms baseline methods in terms of all the evaluation metrics. 
This indicates that HAHT can better encode the history conversations, leverage history conversations to understand the current conversation context and generate more human-like responses.

\subsection{Human Evaluation}

Table \ref{response_huamn_table} summarizes the human evaluation results on the Facebook MSC dataset.
Generally, HAHT outperforms all the baseline methods in terms of all perspectives. This observation is consistent with the automatic evaluation results shown in Table~\ref{all_session_results_table}. In particular, we find that HAHT performs much better than other baselines in terms of history relevancy. This demonstrates that HAHT can better leverage the history conversation sessions and engage the user more in the on-going session with the history memory. HAHT also performs better than other baselines in terms of readability and context relevancy. This indicates that HAHT can better understand the current conversation context with the help of the history memory.

\subsection{Evaluation on Session Openings}
In the MSC task, the session opening is the first conversation turn of the current conversation. According to our observation and the similar observation in~\cite{xu2021beyond}, the opening conversation turn is categorically different from other conversation turns. It typically involves a statement or question that aims to reengage the other speaker based on the known information that has been exchanged in history conversations. Therefore, the performance on the session opening data can further demonstrate the model's capability in understanding and leveraging history conversations. 

We compare all models on these opening responses and show the results in Table~\ref{session_opening_results_table}. We observe that the proposed HAHT model achieves the best performance in terms of most metrics. Especially, when there are 4 history conversations, HAHT outperforms FID-RAG and BlenderBot$_\text{msc}$ by $10.6$\% and $9.1$\% in terms of BLUE-3. This indicates that the proposed HAHT can better leverage conversation history to reengage the user into a new conversation session.

\subsection{Ablation study}
To better understand the effectiveness of each main component of HAHT, we conduct ablation study for HAHT. Specifically, we consider the following variants of HAHT. 
\begin{itemize}
    \item  \textbf{HAHT\textsubscript{w/o HIER}}: In this variant, we do not encode the history conversations hierarchically. Instead, we concatenate all the utterances of history conversations into a long sentence and directly encode it using the transformer encoder.
    
    \item \textbf{HAHT\textsubscript{w/o HIST}}: In this variant, we remove the history encoder from HAHT.
    
    \item \textbf{HAHT\textsubscript{w/o SW}}: In this variant, we remove the switching mechanism from the response generator of HAHT.
\end{itemize}

Table~\ref{ablation_study} summarizes the results achieved by different HAHT variants, in terms of BLEU-2, BLEU-3, and Rouge-L. We note that HAHT outperforms HAHT\textsubscript{w/o HIER}, which indicates that hierarchically encoding the history conversations can help the model reserve more history memory to generate more human-like responses. Moreover, HAHT achieves better performance than HAHT\textsubscript{HIST}. This observation indicates that removing the history encoder causes the most decline in all metrics. This result confirms the necessity to leverage history conversations to understand the current conversation and generate the response. In addition, the performance degradation caused by removing the switching mechanism shows that directly copying words from the history conversation can help the model generate more history-aware responses.

\begin{table}
    \begin{tabular}{p{2cm} p{4.8cm}}
    \hline
    \multicolumn{2}{c}{History Conversation Sessions}\\
    \hline
    {[Session 1]}\\
    ...\\
    User: & ... My dad is a truck driver.\\
    Agent: & ... I am a \textbf{big reader} myself.\\
    User: & ... I like \textbf{working outdoors}.\\
    ...\\
    {[Session 2]}\\
    User: & What book are you currently reading?\\
    Agent: & I'm reading a Stephen King novel. ...\\
    ...\\
    User: &I used to sit in the truck on \textbf{long trips} with my dad, and had nothing to do other than \textbf{read}.\\
    Agent: &Ah, that makes sense. I started reading when my parents were at work. ...\\
    \hline
    \multicolumn{2}{c}{Current Conversations}\\
    \hline
    ...\\
    Agent: & ... What kinds of picnic foods do you like?\\
    User: & I can make sandwiches for us! I also have a very good recipe for homemade lemonade! Do you like lemonade?\\
    \hline
    \multicolumn{2}{c}{Responses from Different Dialogue Systems}\\
    \hline
    BlenderBot: & I love lemonades ! I'm not sure if I could make a picnic for you though. I'd have to make a lot more than just a picnic.\\
    BlenderBot$_{\text{msc}}$: & I love lemonade! I'm not a huge fan of lemonade, but I love lemonade.\\
    FID-RAG: & I love lemonade! I'm sure we can find a good picnic for you.\\
    HAHT (ours): &  I love lemonade! I'm sure we can find a lot of good recipes for sandwiches too. Sandwiches and lemonade are perfect for \textbf{going outdoors} and \textbf{reading books}.\\
    \hline
    \end{tabular}
    \caption{A case study of an MSC with two history conversations. Only important utterances in the history and current conversations are presented. Complete conversations sessions are provided in Appendix \ref{full_case_study}}
    \label{case_study}
\end{table}

\subsection{Case Study} \label{sec:case}

Table~\ref{case_study} shows a case study of the multi-session conversations generated by different models. 
Compared to baseline models, the proposed HAHT model can better leverage history conversations to understand the current conversation context and generate more history-aware responses.
When the user discusses preparing sandwiches and lemonade with the agent, "\textit{I can make sandwiches for us! I also have a very good recipe for homemade lemonade! Do you like lemonade?}", HAHT can remember information mentioned in the history conversations, such as the user likes reading and outdoor activities and it has adopted a book-lover persona before. HAHT can leverage these historical contexts and generate more human-like, context-relevant, and history-aware responses: \textit{``I love lemonade! I’m sure we can find a lot of good recipes for sandwiches too. Sandwiches and lemonade are perfect for going outdoors and reading books.''}.

\section{Conclusion}
In this work, we propose the History-Aware Hierarchical Transformer (HAHT) model for multi-session open-domain dialogue systems. The proposed HAHT model maintains a history memory by hierarchically encoding the history conversation sessions. After that, HAHT uses attention-based encoding to encode the current conversation context together with the history memory and updates the history memory with the current context. In order to explicitly leverage historical information in the responses, HAHT is designed with a history-aware response generator which can switch between a generic vocabulary and a history-aware vocabulary. Experimental results obtained under both normal and session opening MSC settings demonstrate that HAHT performs better in conducting MSC and generates more human-like, context-relevant, and history-aware responses than state-of-the-art models.

\section{Limitations}
One limitation of this work is that HAHT has only been evaluated on one dataset. However, to the best of best of our knowledge, Facebook MSC is, by far, the only large-scale multi-session conversation dataset available. 
Nevertheless, our proposed model consistently outperforms baseline models on conversations with different numbers of history sessions in Facebook MSC.

A potential solution to this limitation is to construct more MSC datasets in open-domain or in specific-domain that may benefit from the awareness of history conversations, \eg conversational recommendation or automatic medical assistants. 

\section{Acknowledgments}
This research is supported, in part, by Alibaba Group through the Alibaba Innovative Research (AIR) Program and the Alibaba-NTU Singapore Joint Research Institute (AN-GC-2021-04), by the National Research Foundation (NRF) Investigatorship Programme (NRF-NRFI05-2019-0002), and by the NRF Fellowship (NRF-NRFF13-2021-0006). Any opinions, findings and conclusions or recommendations expressed in this material are those of the authors and do not reflect the views of the funding agencies.

\bibliography{anthology,custom}

\begin{thebibliography}{34}
\expandafter\ifx\csname natexlab\endcsname\relax\def\natexlab#1{#1}\fi

\bibitem[{Adiwardana et~al.(2020)Adiwardana, Luong, So, Hall, Fiedel,
  Thoppilan, Yang, Kulshreshtha, Nemade, Lu, and Le}]{adiwardana2020towards}
Daniel Adiwardana, Minh-Thang Luong, David~R So, Jamie Hall, Noah Fiedel, Romal
  Thoppilan, Zi~Yang, Apoorv Kulshreshtha, Gaurav Nemade, Yifeng Lu, and
  Quoc~V. Le. 2020.
\newblock \href {https://arxiv.org/abs/2001.09977} {Towards a human-like
  open-domain chatbot}.
\newblock \emph{arXiv preprint arXiv:2001.09977}.

\bibitem[{Clark et~al.(2019)Clark, Pantidi, Cooney, Doyle, Garaialde, Edwards,
  Spillane, Gilmartin, Murad, Munteanu, Wade, and Cowan}]{clark2019makes}
Leigh Clark, Nadia Pantidi, Orla Cooney, Philip Doyle, Diego Garaialde, Justin
  Edwards, Brendan Spillane, Emer Gilmartin, Christine Murad, Cosmin Munteanu,
  Vincent Wade, and Benjamin~R. Cowan. 2019.
\newblock \href {https://doi.org/10.1145/3290605.3300705} {{What makes a good
  conversation? Challenges in designing truly conversational agents}}.
\newblock In \emph{Proceedings of the 2019 CHI Conference on Human Factors in
  Computing Systems}, pages 1--12, New York, NY, USA. Association for Computing
  Machinery.

\bibitem[{Dinan et~al.(2019)Dinan, Roller, Shuster, Fan, Auli, and
  Weston}]{dinan2018wizard}
Emily Dinan, Stephen Roller, Kurt Shuster, Angela Fan, Michael Auli, and Jason
  Weston. 2019.
\newblock \href {https://openreview.net/forum?id=r1l73iRqKm} {{Wizard of
  Wikipedia: Knowledge-powered conversational agents}}.
\newblock In \emph{Proceedings of 7th International Conference on Learning
  Representations}.

\bibitem[{Fleiss and Cohen(1973)}]{kappa}
Joseph~L. Fleiss and Jacob Cohen. 1973.
\newblock \href {https://doi.org/10.1177/001316447303300309} {The equivalence
  of weighted kappa and the intraclass correlation coefficient as measures of
  reliability}.
\newblock \emph{Educational and Psychological Measurement}, 33(3):613--619.

\bibitem[{Gu et~al.(2016)Gu, Lu, Li, and Li}]{gu-etal-2016-incorporating}
Jiatao Gu, Zhengdong Lu, Hang Li, and Victor~O.K. Li. 2016.
\newblock \href {https://doi.org/10.18653/v1/P16-1154} {Incorporating copying
  mechanism in sequence-to-sequence learning}.
\newblock In \emph{Proceedings of the 54th Annual Meeting of the Association
  for Computational Linguistics (Volume 1: Long Papers)}, pages 1631--1640,
  Berlin, Germany. Association for Computational Linguistics.

\bibitem[{Gulcehre et~al.(2016)Gulcehre, Ahn, Nallapati, Zhou, and
  Bengio}]{gulcehre-etal-2016-pointing}
Caglar Gulcehre, Sungjin Ahn, Ramesh Nallapati, Bowen Zhou, and Yoshua Bengio.
  2016.
\newblock \href {https://doi.org/10.18653/v1/P16-1014} {Pointing the unknown
  words}.
\newblock In \emph{Proceedings of the 54th Annual Meeting of the Association
  for Computational Linguistics (Volume 1: Long Papers)}, pages 140--149,
  Berlin, Germany. Association for Computational Linguistics.

\bibitem[{Higashinaka et~al.(2014)Higashinaka, Imamura, Meguro, Miyazaki,
  Kobayashi, Sugiyama, Hirano, Makino, and
  Matsuo}]{higashinaka-etal-2014-towards}
Ryuichiro Higashinaka, Kenji Imamura, Toyomi Meguro, Chiaki Miyazaki, Nozomi
  Kobayashi, Hiroaki Sugiyama, Toru Hirano, Toshiro Makino, and Yoshihiro
  Matsuo. 2014.
\newblock \href {https://aclanthology.org/C14-1088} {Towards an open-domain
  conversational system fully based on natural language processing}.
\newblock In \emph{Proceedings of {COLING} 2014, the 25th International
  Conference on Computational Linguistics: Technical Papers}, pages 928--939,
  Dublin, Ireland. Dublin City University and Association for Computational
  Linguistics.

\bibitem[{Huang et~al.(2020)Huang, Zhu, and Gao}]{huang2020challenges}
Minlie Huang, Xiaoyan Zhu, and Jianfeng Gao. 2020.
\newblock Challenges in building intelligent open-domain dialog systems.
\newblock \emph{ACM Transactions on Information Systems (TOIS)}, 38(3):1--32.

\bibitem[{Izacard and Grave(2021)}]{izacard-grave-2021-leveraging}
Gautier Izacard and Edouard Grave. 2021.
\newblock \href {https://doi.org/10.18653/v1/2021.eacl-main.74} {Leveraging
  passage retrieval with generative models for open domain question answering}.
\newblock In \emph{Proceedings of the 16th Conference of the European Chapter
  of the Association for Computational Linguistics: Main Volume}, pages
  874--880, Online. Association for Computational Linguistics.

\bibitem[{Kingma and Ba(2014)}]{kingma2014adam}
Diederik~P Kingma and Jimmy Ba. 2014.
\newblock \href {https://arxiv.org/abs/1412.6980} {Adam: A method for
  stochastic optimization}.
\newblock \emph{arXiv preprint arXiv:1412.6980}.

\bibitem[{Lewis et~al.(2020)Lewis, Perez, Piktus, Petroni, Karpukhin, Goyal,
  K{\"u}ttler, Lewis, Yih, Rockt{\"a}schel, Riedel, and
  Kiela}]{lewis2020retrieval}
Patrick Lewis, Ethan Perez, Aleksandra Piktus, Fabio Petroni, Vladimir
  Karpukhin, Naman Goyal, Heinrich K{\"u}ttler, Mike Lewis, Wen-tau Yih, Tim
  Rockt{\"a}schel, Sebastian Riedel, and Douwe Kiela. 2020.
\newblock \href
  {https://proceedings.neurips.cc/paper/2020/hash/6b493230205f780e1bc26945df7481e5-Abstract.html}
  {Retrieval-augmented generation for knowledge-intensive nlp tasks}.
\newblock In \emph{Advances in Neural Information Processing Systems},
  volume~33, pages 9459--9474. Curran Associates, Inc.

\bibitem[{Li et~al.(2016)Li, Galley, Brockett, Gao, and
  Dolan}]{li-etal-2016-diversity}
Jiwei Li, Michel Galley, Chris Brockett, Jianfeng Gao, and Bill Dolan. 2016.
\newblock \href {https://doi.org/10.18653/v1/N16-1014} {A diversity-promoting
  objective function for neural conversation models}.
\newblock In \emph{Proceedings of the 2016 Conference of the North {A}merican
  Chapter of the Association for Computational Linguistics: Human Language
  Technologies}, pages 110--119, San Diego, California. Association for
  Computational Linguistics.

\bibitem[{Li et~al.(2017)Li, Su, Shen, Li, Cao, and
  Niu}]{li-etal-2017-dailydialog}
Yanran Li, Hui Su, Xiaoyu Shen, Wenjie Li, Ziqiang Cao, and Shuzi Niu. 2017.
\newblock \href {https://aclanthology.org/I17-1099} {{D}aily{D}ialog: A
  manually labelled multi-turn dialogue dataset}.
\newblock In \emph{Proceedings of the Eighth International Joint Conference on
  Natural Language Processing (Volume 1: Long Papers)}, pages 986--995, Taipei,
  Taiwan. Asian Federation of Natural Language Processing.

\bibitem[{Lin and Och(2004)}]{lin-och-2004-automatic}
Chin-Yew Lin and Franz~Josef Och. 2004.
\newblock \href {https://doi.org/10.3115/1218955.1219032} {Automatic evaluation
  of machine translation quality using longest common subsequence and
  skip-bigram statistics}.
\newblock In \emph{Proceedings of the 42nd Annual Meeting of the Association
  for Computational Linguistics ({ACL}-04)}, pages 605--612, Barcelona, Spain.

\bibitem[{Lin et~al.(2017)Lin, Feng, Santos, Yu, Xiang, Zhou, and
  Bengio}]{lin2017iclr}
Zhouhan Lin, Minwei Feng, Cicero Nogueira~dos Santos, Mo~Yu, Bing Xiang, Bowen
  Zhou, and Yoshua Bengio. 2017.
\newblock \href {https://openreview.net/forum?id=BJC\_jUqxe} {A structured
  self-attentive sentence embedding}.
\newblock In \emph{Proceedings of 5th International Conference on Learning
  Representations}.

\bibitem[{Papineni et~al.(2002)Papineni, Roukos, Ward, and
  Zhu}]{papineni-etal-2002-bleu}
Kishore Papineni, Salim Roukos, Todd Ward, and Wei-Jing Zhu. 2002.
\newblock \href {https://doi.org/10.3115/1073083.1073135} {{B}leu: a method for
  automatic evaluation of machine translation}.
\newblock In \emph{Proceedings of the 40th Annual Meeting of the Association
  for Computational Linguistics}, pages 311--318, Philadelphia, Pennsylvania,
  USA. Association for Computational Linguistics.

\bibitem[{Platonov et~al.(2020)Platonov, Schubert, Kane, and
  Gindi}]{platonov-etal-2020-spoken}
Georgiy Platonov, Lenhart Schubert, Benjamin Kane, and Aaron Gindi. 2020.
\newblock \href {https://aclanthology.org/2020.sigdial-1.16} {A spoken dialogue
  system for spatial question answering in a physical blocks world}.
\newblock In \emph{Proceedings of the 21th Annual Meeting of the Special
  Interest Group on Discourse and Dialogue}, pages 128--131, 1st virtual
  meeting. Association for Computational Linguistics.

\bibitem[{Qian et~al.(2021)Qian, Dou, Zhu, Ma, and Wen}]{qian-cikm-2021}
Hongjin Qian, Zhicheng Dou, Yutao Zhu, Yueyuan Ma, and Ji-Rong Wen. 2021.
\newblock \href {https://doi.org/10.1145/3459637.3482269} {Learning implicit
  user profile for personalized retrieval-based chatbot}.
\newblock In \emph{Proceedings of the 30th ACM International Conference on
  Information and Knowledge Management}, page 1467–1477, New York, NY, USA.
  Association for Computing Machinery.

\bibitem[{Ritter et~al.(2011)Ritter, Cherry, and Dolan}]{ritter-etal-2011-data}
Alan Ritter, Colin Cherry, and William~B. Dolan. 2011.
\newblock \href {https://aclanthology.org/D11-1054} {Data-driven response
  generation in social media}.
\newblock In \emph{Proceedings of the 2011 Conference on Empirical Methods in
  Natural Language Processing}, pages 583--593, Edinburgh, Scotland, UK.
  Association for Computational Linguistics.

\bibitem[{Roller et~al.(2020)Roller, Boureau, Weston, Bordes, Dinan, Fan,
  Gunning, Ju, Li, Poff, Ringshia, Shuster, Smith, Szlam, Urbanek, and
  Williamson}]{roller2020open}
Stephen Roller, Y-Lan Boureau, Jason Weston, Antoine Bordes, Emily Dinan,
  Angela Fan, David Gunning, Da~Ju, Margaret Li, Spencer Poff, Pratik Ringshia,
  Kurt Shuster, Eric~Michael Smith, Arthur Szlam, Jack Urbanek, and Mary
  Williamson. 2020.
\newblock \href {https://arxiv.org/abs/2006.12442} {{Open-domain conversational
  agents: Current progress, open problems, and future directions}}.
\newblock \emph{arXiv preprint arXiv:2006.12442}.

\bibitem[{Roller et~al.(2021)Roller, Dinan, Goyal, Ju, Williamson, Liu, Xu,
  Ott, Smith, Boureau, and Weston}]{roller-etal-2021-recipes}
Stephen Roller, Emily Dinan, Naman Goyal, Da~Ju, Mary Williamson, Yinhan Liu,
  Jing Xu, Myle Ott, Eric~Michael Smith, Y-Lan Boureau, and Jason Weston. 2021.
\newblock \href {https://doi.org/10.18653/v1/2021.eacl-main.24} {Recipes for
  building an open-domain chatbot}.
\newblock In \emph{Proceedings of the 16th Conference of the European Chapter
  of the Association for Computational Linguistics: Main Volume}, pages
  300--325, Online. Association for Computational Linguistics.

\bibitem[{Serban et~al.(2016)Serban, Sordoni, Bengio, Courville, and
  Pineau}]{serban2016building}
Iulian~V. Serban, Alessandro Sordoni, Yoshua Bengio, Aaron Courville, and
  Joelle Pineau. 2016.
\newblock \href {https://dl.acm.org/doi/abs/10.5555/3504035.3504723} {Building
  end-to-end dialogue systems using generative hierarchical neural network
  models}.
\newblock In \emph{Proceedings of the Thirtieth AAAI Conference on Artificial
  Intelligence}, page 3776–3783. AAAI Press.

\bibitem[{Serban et~al.(2017)Serban, Sordoni, Lowe, Charlin, Pineau, Courville,
  and Bengio}]{VHRED2016}
Iulian~Vlad Serban, Alessandro Sordoni, Ryan Lowe, Laurent Charlin, Joelle
  Pineau, Aaron Courville, and Yoshua Bengio. 2017.
\newblock \href {https://dl.acm.org/doi/10.5555/3298023.3298047} {A
  hierarchical latent variable encoder-decoder model for generating dialogues}.
\newblock In \emph{Proceedings of the Thirty-First AAAI Conference on
  Artificial Intelligence}, pages 3295–--3301. AAAI Press.

\bibitem[{Shum et~al.(2018)Shum, He, and Li}]{xiaoice-2018}
Heung-Yeung Shum, Xiaodong He, and Di~Li. 2018.
\newblock \href {https://arxiv.org/abs/1801.01957} {{From Eliza to XiaoIce:
  Challenges and opportunities with social chatbots}}.
\newblock \emph{arXiv preprint arXiv:1801.01957}.

\bibitem[{Shuster et~al.(2021)Shuster, Poff, Chen, Kiela, and
  Weston}]{shuster-etal-2021-retrieval-augmentation}
Kurt Shuster, Spencer Poff, Moya Chen, Douwe Kiela, and Jason Weston. 2021.
\newblock \href {https://doi.org/10.18653/v1/2021.findings-emnlp.320}
  {Retrieval augmentation reduces hallucination in conversation}.
\newblock In \emph{Findings of the Association for Computational Linguistics:
  EMNLP 2021}, pages 3784--3803, Punta Cana, Dominican Republic. Association
  for Computational Linguistics.

\bibitem[{Vaswani et~al.(2017)Vaswani, Shazeer, Parmar, Uszkoreit, Jones,
  Gomez, Kaiser, and Polosukhin}]{vaswani2017attention}
Ashish Vaswani, Noam Shazeer, Niki Parmar, Jakob Uszkoreit, Llion Jones,
  Aidan~N Gomez, {\L}ukasz Kaiser, and Illia Polosukhin. 2017.
\newblock \href
  {https://papers.nips.cc/paper/2017/hash/3f5ee243547dee91fbd053c1c4a845aa-Abstract.html}
  {Attention is all you need}.
\newblock In \emph{Advances in Neural Information Processing Systems},
  volume~30, pages 5998--6008. Curran Associates, Inc.

\bibitem[{Vinyals and Le(2015)}]{vinyals2015neural}
Oriol Vinyals and Quoc Le. 2015.
\newblock \href {https://arxiv.org/abs/1506.05869} {A neural conversational
  model}.
\newblock \emph{arXiv preprint arXiv:1506.05869}.

\bibitem[{Xu et~al.(2022)Xu, Szlam, and Weston}]{xu2021beyond}
Jing Xu, Arthur Szlam, and Jason Weston. 2022.
\newblock \href {https://doi.org/10.18653/v1/2022.acl-long.356} {{Beyond
  goldfish memory: Long-term open-domain conversation}}.
\newblock In \emph{Proceedings of the 60th Annual Meeting of the Association
  for Computational Linguistics (Volume 1: Long Papers)}, pages 5180--5197,
  Dublin, Ireland. Association for Computational Linguistics.

\bibitem[{Yuan et~al.(2019)Yuan, Zhou, Li, Lv, Zhu, Han, and
  Hu}]{yuan-etal-2019-multi}
Chunyuan Yuan, Wei Zhou, Mingming Li, Shangwen Lv, Fuqing Zhu, Jizhong Han, and
  Songlin Hu. 2019.
\newblock \href {https://doi.org/10.18653/v1/D19-1011} {Multi-hop selector
  network for multi-turn response selection in retrieval-based chatbots}.
\newblock In \emph{Proceedings of the 2019 Conference on Empirical Methods in
  Natural Language Processing and the 9th International Joint Conference on
  Natural Language Processing (EMNLP-IJCNLP)}, pages 111--120, Hong Kong,
  China. Association for Computational Linguistics.

\bibitem[{Zhang et~al.(2018)Zhang, Dinan, Urbanek, Szlam, Kiela, and
  Weston}]{zhang-etal-2018-personachat}
Saizheng Zhang, Emily Dinan, Jack Urbanek, Arthur Szlam, Douwe Kiela, and Jason
  Weston. 2018.
\newblock \href {https://doi.org/10.18653/v1/P18-1205} {Personalizing dialogue
  agents: {I} have a dog, do you have pets too?}
\newblock In \emph{Proceedings of the 56th Annual Meeting of the Association
  for Computational Linguistics (Volume 1: Long Papers)}, pages 2204--2213,
  Melbourne, Australia. Association for Computational Linguistics.

\bibitem[{Zhang et~al.(2020)Zhang, Sun, Galley, Chen, Brockett, Gao, Gao, Liu,
  and Dolan}]{zhang-etal-2020-dialogpt}
Yizhe Zhang, Siqi Sun, Michel Galley, Yen-Chun Chen, Chris Brockett, Xiang Gao,
  Jianfeng Gao, Jingjing Liu, and Bill Dolan. 2020.
\newblock \href {https://doi.org/10.18653/v1/2020.acl-demos.30} {{DIALOGPT} :
  Large-scale generative pre-training for conversational response generation}.
\newblock In \emph{Proceedings of the 58th Annual Meeting of the Association
  for Computational Linguistics: System Demonstrations}, pages 270--278,
  Online. Association for Computational Linguistics.

\bibitem[{Zhong et~al.(2020)Zhong, Zhang, Wang, Liu, and
  Miao}]{zhong-etal-2020-towards}
Peixiang Zhong, Chen Zhang, Hao Wang, Yong Liu, and Chunyan Miao. 2020.
\newblock \href {https://doi.org/10.18653/v1/2020.emnlp-main.531} {Towards
  persona-based empathetic conversational models}.
\newblock In \emph{Proceedings of the 2020 Conference on Empirical Methods in
  Natural Language Processing (EMNLP)}, pages 6556--6566, Online. Association
  for Computational Linguistics.

\bibitem[{Zhou et~al.(2016)Zhou, Dong, Wu, Zhao, Yu, Tian, Liu, and
  Yan}]{zhou-etal-2016-multi}
Xiangyang Zhou, Daxiang Dong, Hua Wu, Shiqi Zhao, Dianhai Yu, Hao Tian, Xuan
  Liu, and Rui Yan. 2016.
\newblock \href {https://doi.org/10.18653/v1/D16-1036} {Multi-view response
  selection for human-computer conversation}.
\newblock In \emph{Proceedings of the 2016 Conference on Empirical Methods in
  Natural Language Processing}, pages 372--381, Austin, Texas. Association for
  Computational Linguistics.

\bibitem[{Zhu et~al.(2021)Zhu, Nie, Zhou, Du, and Dou}]{retrieval-zhou-2021}
Yutao Zhu, Jian-Yun Nie, Kun Zhou, Pan Du, and Zhicheng Dou. 2021.
\newblock \href {https://doi.org/10.1007/978-3-030-72113-8_50} {Content
  selection network for document-grounded retrieval-based chatbots}.
\newblock In \emph{Advances in Information Retrieval: 43rd European Conference
  on IR Research, ECIR}, page 755–769, Berlin, Heidelberg. Springer-Verlag.

\end{thebibliography}
\bibliographystyle{acl_natbib}

\appendix
\label{sec:appendix}

\section{Appendix}
\subsection{Complete Case Study}
\label{full_case_study}
Table \ref{complete_case_study} shows the complete content of the case study in section~\ref{sec:case}. The conversations were conducted by two crowd workers. For training purposes, the utterances from the first crowd worker are denoted as user utterances, whereas those from the second crowd worker are denoted as agent utterances. An MSC model aims to generate responses that are as close as possible to the second crowd worker's utterances.

In history session 1, the user and the agent briefly talk about themselves, such as, their hobbies, and the job of their parents. In history session 2, they talk about the book that they are reading and some memories about reading books. In the current conversation, their discussions vary from the book 1984 to the plan of hiking and picnic. When the user discusses preparing sandwiches and lemonade with the agent, "\textit{I can make sandwiches for us! I also have a very good recipe for homemade lemonade! Do you like lemonade?}", HAHT can remember information mentioned in the history conversations, such as the user likes reading and outdoor activities and it has adopted a book-lover persona before. HAHT can leverage these historical contexts and generate more human-like, context-relevant, and history-aware responses: \textit{``I love lemonade! I’m sure we can find a lot of good recipes for sandwiches too. Sandwiches and lemonade are perfect for going outdoors and reading books.''}.

\onecolumn
\begin{longtable}{p{2cm} p{13.1cm}}
    \hline
    \multicolumn{2}{c}{History Conversation Sessions}\\
    \hline
    [Session 1]\\
    User: & Hello! Tell me something about yourself!\\
    Agent: & I like animals and have a kind heart.\\
    User: & That's really nice :) it's cool to find someone with a nice heart.\\ 
    Agent: & Yes, having a doctor mom and fireman dad has made me very kind in general.\\
    User: & That's great! My parents are a bit different, my dad is a truck driver.\\
    Agent: & Does he ever read while driving? I am a big reader myself.\\
    User: & That's dangerous. We're more of an active family though, I like working outdoors.\\
    Agent: & I prefer to sit on the couch with my animal friends and read.\\
    User: & Nice! I also play basketball often. My outdoor genes come from my canadian mother.\\
    Agent: & I sometimes like to go hiking with my animal friends as well.\\
    User: & Oh nice :) hiking is tough because it takes me longer to walk, since I'm short.\\
    Agent: & I'm really kind, so I wouldn't make fun of you if you went hiking with me.\\
    {[Session 2]}\\
    User: & What book are you currently reading?\\
    Agent: & I'm reading a Stephen King novel. Are you reading anything? What's your favorite genre of book?\\
    User: & Oh nice, I like Stephen King too. Which of his novels are you reading? I like dystopian fiction, currently reading Margaret Atwood, supporting Canadian authors lol\\
    Agent: & I'm reading his newest novel, Later. Oh, I've seen a TV adaptation of one of Margaret Atwood's books!\\
    User: & Is it Handmaid's tale? I love that show\\
    Agent: & Yes, me too. It was dark, but thought provoking. Do you recommend anything else like it?\\
    User: & After I read Handmaid's Tale I read 1984, which was similar. I read it in highschool, but couldn't truly appreciate it back then.\\
    Agent: & 1984. That's written by George Orwell?\\
    User: & Yup, he also wrote Animal Farm, which is also very dark and thought provoking, and you might like it since it's about animals!\\
    Agent: &Yeah, I do like dark and thought provoking movies. Bonus that it's about animals. I'll try giving it a shot. You seem to know a lot of literature.\\
    User: &I used to sit in the truck on long trips with my dad, and had nothing to do other than read.\\
    Agent: &Ah, that makes sense. I started reading when my parents were at work. My mom was working a lot, and sometimes she brought me to the hospital where I'd read at her desk. What does your mom do for work?\\
    \hline
    \multicolumn{2}{c}{Current Conversations}\\
    \hline
    User: & I looked into the 1984 book we were speaking about before.\\
    Agent: & oh ya? did you start reading it yet?\\
    User: & Yeah! It was much better this time around I definitely understood all of the nuances now that I am older!\\
    Agent: & I always thought it was a little bit weird to have us read it in high school. definitely more to appreciate to read as an adult i think. have you picked out the next book that you want to read yet?\\
    User: & I'm not sure what I want to read next! I can be very indecisive. Do you have a good book to recommend?\\
    Agent: & hmm.....i read a book once call 1984 which is a very surreal type book, which sounds like what you like! warning though, it is very long. i really think it might be up your alley if you can commit to that long of a book\\
    User: & I definitely have a lot of free time right so I will take a look at it!\\
    Agent: & then that is perfect! it should keep you busy for quite a long time. we should schedule a hike together sometime to balance out all the reading with some exercise too. would you be interested?\\
    User: & As long as we go slow so I can keep up! Where do you usually hike?\\
    Agent: & we can go as fast or slow as you like. it's not as much about the speed for me anyway. i really like to enjoy and take in the sights and the fresh air. then great, i will find a good hike for us, and maybe we could plan a picnic too. what kinds of picnic foods do you like?\\
    User: & I can make sandwiches for us! I also have a very good recipe for homemade lemonade! Do you like lemonade?\\
    \hline
    \multicolumn{2}{c}{Responses from Different Dialogue Systems}\\
    \hline
    BlenderBot: & I love lemonades ! I'm not sure if I could make a picnic for you though. I'd have to make a lot more than just a picnic.\\
    BlenderBot$_{\text{msc}}$: & I love lemonade! I'm not a huge fan of lemonade, but I love lemonade.\\
    FID-RAG: & I love lemonade! I'm sure we can find a good picnic for you.\\
    HAHT (ours): &  I love lemonade! I'm sure we can find a lot of good recipes for sandwiches too. Sandwiches and lemonade are perfect for going outdoors and reading books.\\
    \hline
    \caption{Case study of different model replies on a two-session conversations.}\\
    \label{complete_case_study}
\end{longtable}
\end{document}